\documentclass{article}

\PassOptionsToPackage{numbers, compress}{natbib}

\usepackage[preprint]{nips_2018}

\usepackage[utf8]{inputenc} 
\usepackage[T1]{fontenc}    
\usepackage{hyperref}       
\usepackage{url}            
\usepackage{booktabs}       
\usepackage{amsfonts}       
\usepackage{nicefrac}       
\usepackage{microtype}      
\usepackage{graphicx}
\usepackage{algorithm}
\usepackage{algpseudocode}
\usepackage{amsmath,amsthm,amssymb,amsfonts}
\usepackage{epstopdf}
\usepackage{caption}
\usepackage{subcaption}

\title{Abstraction Learning}

\author{
  Fei Deng \\
  \texttt{fei.deng@rutgers.edu} \\
  \And Jinsheng Ren\\
   \texttt{rjs17@mails.tsinghua.edu.cn}
  \AND Feng Chen\\
  \texttt{chenfeng@mail.tsinghua.edu.cn}
}

\begin{document}
	\maketitle
	\begin{abstract}
	There has been a gap between artificial intelligence and human intelligence. In this paper, we identify three key elements forming human intelligence, and suggest that abstraction learning combines these elements and is thus a way to bridge the gap. Prior researches in artificial intelligence either specify abstraction by human experts, or take abstraction as qualitative explanation for the model. This paper aims to learn abstraction directly. We tackle three main challenges: representation, objective function, and learning algorithm. Specifically, we propose a partition structure that contains pre-allocated abstraction neurons; we formulate abstraction learning as a constrained optimization problem, which integrates abstraction properties; we develop a network evolution algorithm to solve this problem. This complete framework is named \emph{\textbf{ONE}} (Optimization via Network Evolution). In our experiments on MNIST, ONE shows elementary human-like intelligence, including low energy consumption, knowledge sharing, and lifelong learning.
	\end{abstract}
\vspace{-1 em}
\section{Introduction}
\vspace{-0.8 em}
Human intelligence has long been the goal of artificial intelligence. Humans can effectively solve multi-task learning, lifelong learning, transfer learning, few-shot learning, generalization, exploration, prediction, and decision. AI algorithms, on the other hand, still struggle with these problems\cite{legg2007universal}. For example, supervised learning based on neural networks typically requires large training datasets, and suffers from catastrophic forgetting\cite{kirkpatrick2017overcoming,rusu2016progressive,goodfellow2013empirical}; most reinforcement learning algorithms\cite{parisotto2015actor,schaul2015universal,rusu2015policy} specialize in a single task through huge amounts of trial and error, and are difficult to generalize to new tasks.

So how is human intelligence formed? This remains an open question, and we hypothesize that there are at least three key elements. 1) \emph{Intrinsic motivation}. Humans are intrinsically motivated to truly understand the world. This understanding is crucial to intelligence because it applies to and is required by all tasks that humans will be faced with. 2) \emph{Unified network}. Humans use a single unified network – the brain – to tackle all tasks. This allows knowledge to be accumulated and shared across tasks, and thus increase the level of intelligence over time. 3) \emph{Limited complexity}. Humans face several constraints on complexity. They have limited time and data to learn each task; they have limited space to encode and pass on knowledge; the brain has limited energy to operate. These constraints force humans to develop intelligence and find efficient solutions.

Interestingly, we find that abstraction learning is a way to combine these elements, and thus key to achieving human-like intelligence. Here, we define abstractions as the set of concepts and laws that the world is built upon. By this definition, learning abstractions is equivalent to discovering a universal model of the world and using it to interpret observations. This satisfies intrinsic motivation, and has two implications. First, because only a single world model is yielded by abstraction learning, it can be maintained in a unified network. This model can then be shared and improved across all tasks, facilitating knowledge transfer and generalization. Second, because abstraction learning enables interpretation of complex observations as concise concepts and laws, task models can be greatly simplified. Consequently, fewer samples are required to learn each task, and less energy is needed to perform each task. 

While there is biological evidence suggesting the existence of abstractions in the brain (e.g. the grandmother cell\cite{gross2002genealogy,clark2000theory,knorkski1967integrative}), abstraction learning remains largely unexplored in AI research. One of the reasons why abstraction learning is harder than other machine learning tasks such as classification is because it is difficult to establish a clear objective, or target for training\cite{bengio2013representation}. Two broadly related areas are probabilistic graphical models and neural networks. Probabilistic graphical models\cite{roweis1998algorithms,olshausen1996emergence,smolensky1986information} use abstractions to define random variables and graph structures. These abstractions serve to decompose the joint probability distribution and simplify computation, but they are given by human experts rather than learned by the model itself. Neural networks, on the other hand, aim to learn effective representations for specific tasks. Taking auto-encoder\cite{yann1987modeles,bourlard1988auto,hinton1994autoencoders} as an example, one can learn a set of representation for data by unsupervised learning, typically aiming for dimensionality reduction. The process of closely matching the output with the original data enforces the auto-encoder to learn the abstraction. Through visualization\cite{zeiler2014visualizing,simonyan2013deep,springenberg2014striving}, these artificial neurons can be qualitatively regarded as representing abstract concepts. In lifelong learning, knowledge shared between tasks was considered as abstractions in recent works, such as using a regularizer that prevents the parameters from drastic changes in their values\cite{kirkpatrick2017overcoming} or blocking any changes to the old task parameters\cite{rusu2016progressive}. However, these abstractions are not directly learned, but emerge as a by-product of minimizing task losses. Therefore, the quality of these abstractions cannot be guaranteed, depriving neural networks of the benefits that real abstractions would offer.

In this paper, we aim to learn abstractions directly. This raises at least three challenges. First, while abstractions may be straightforward to express in natural language, they are hard to specify in the language of neurons and synapses. Second, because abstractions are like hidden variables that do not appear in the training data, there is no simple objective function that distinguishes between right and wrong abstractions. Third, even if we find such an objective function, which would probably be non-differentiable, how it should influence the network structure remains unclear. To overcome these challenges, we propose a novel framework named ONE that formulates abstraction learning as Optimization via Network Evolution. ONE incorporates three levels of innovations:

\quad  $\bullet $ \textbf{Partition structure prior, structure prior, with pre-allocated abstraction neurons.} These abstraction neurons accumulate abstractions, and separate the network into a task-agnostic cognition part and task-specific decision parts. The cognition part generates abstractions, and the decision parts select abstractions. Task losses do not modify abstractions, thus improving reusability and avoiding catastrophic forgetting.

\quad  $\bullet $ \textbf{Constrained optimization formulation, integrating three abstraction properties.} 1) Variety. Abstractions should cover the various concepts and laws behind observations. 2) Simplicity. Each observation must be described succinctly by only a few abstractions, thus simplifying task-specific decision making. 3) Effectiveness. Each abstraction should be effective for multiple tasks, enhancing task performance and knowledge sharing.

\quad  $\bullet $ \textbf{Network evolution algorithm, producing and improving abstraction structures.} The constrained optimization problem for abstraction learning involves optimizing the network structure, which is beyond the capabilities of standard gradient-based methods. To solve this problem, we introduce connection growth to search through the structure space, local competition to improve search efficiency, and use the objective function to guide search direction.

Through extensive experiments on the MNIST dataset, we demonstrate that ONE successfully converges and learns abstractions that accomplish different tasks. Importantly, ONE shows elementary human-like intelligence in three aspects. First, ONE performs tasks based on simple abstractions, and thus activate only a small proportion of the whole network for each task. This leads to better generalization and less energy consumption. Second, when faced with new tasks, ONE is able to generate new abstractions. These abstractions accumulate, so as ONE learns more tasks, it makes more use of existing abstractions and generates fewer new ones. This facilitates task transfer and boosts learning speed. Third, ONE does not forget, and can thus learn continually.

\section{Partition structure prior}
\begin{figure}[t]
	\centering
	\includegraphics[width=0.9\linewidth]{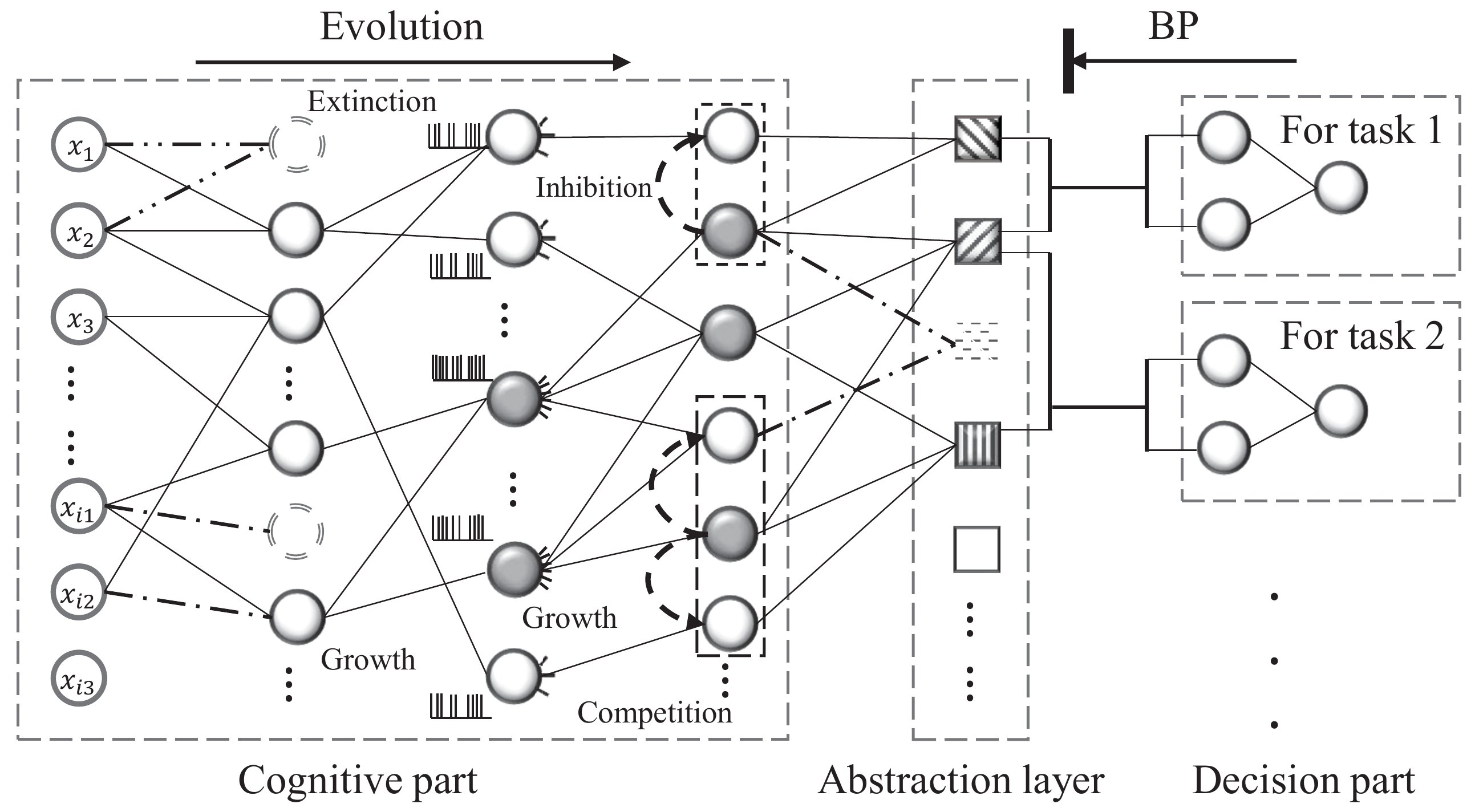}
	\caption[Partition structure of ONE model]{Partition structure of ONE model: ONE performs structure search in the cognitive part and parameters learning in the decision part, with an abstraction locked layer in the middle of them. In the cognitive part, the mechanisms of growth, extinction, and competition promote the production of excellent structures (abstractions), where the dashed line represents death, the number of pulses denotes the firing intensity of neurons and the dark units are winners. The mechanisms of each layer is applied to all layers of the cognitive part. }
	\label{fig:Partition structure of ONE model}
	\vspace{-1.75 em}
\end{figure}

\vspace{-0.8 em}
To our knowledge, although abstraction has been proposed in a large number of literature\cite{hinton1986learning,bengio2012deep,krizhevsky2012imagenet,collobert2011natural}, the current research remains basically at the level of giving general concept and description of abstractions, which we refer to as virtual abstractions. We consider that the process human beings realizing basic intellectual activities can be divided into the cognitive process for concepts and laws and the decision making process related to tasks. Because of the lack of physical meanings of virtual abstractions, it is difficult to draw a clear line between the parts of cognition and decision-making, which makes the channels of knowledge sharing among tasks obscure and the decision-making process of a single task complicated. Therefore, in this section, we propose a specific form of entity abstraction and a partition structure based on the abstraction locked layer (see Figure  \ref{fig:Partition structure of ONE model}).

Moreover, the structures (abstractions) activated by limited amounts of tasks cannot reach the level of understanding the whole world, abstractions needs to be continuously generated, accumulated, and reused by different tasks. In other words, each task makes full use of the existing abstractions, and only generates necessary abstractions that are conducive to the new task. This promotes the accumulation and reuse of abstractions and avoids forgetting knowledge acquired in the past.

The whole network can be seen as a feedforward neural network with a fixed number of neurons, which has been divided into the cognition and decision parts by an abstraction locked layer (see Figure  \ref{fig:Partition structure of ONE model}). The abstraction layer separates neurons into two types: task-agnostic and task-specific. Each neuron in the abstraction layer is the output of a subnetwork composed of task-agnostic neurons. These subnetworks can overlap and take any form, in addition, they should be sparse and hierarchical. The task-specific neurons are organized into non-overlapping groups. The neurons within each group form fully connected layers and are intended for a specific task. Each group connects sparsely with the abstraction layer, selecting the relevant abstractions for each task. Inspired by biology\cite{dayan2001theoretical}, there are three mechanisms that promote the production of excellent abstractions: 1) \textbf{Growth:} This a way of exploring the structure space. The more extensively we explore, the better structure we expect to find. 2) \textbf{Competition:} It promotes the variety and simplicity of abstractions and prunes unpromising structures on the fly. 3) \textbf{Extinction:} It aims to optimize the abstractions for task effectiveness.

Formally, suppose there are $K$ tasks in total. Each task is associated with a dataset $\mathcal{D}^{\left(k\right)}$ and a loss function $\mathcal{L}^{\left(k\right)}$, and requires to learn a mapping $f^{\left(k\right)}:\mathcal{X}^{\left(k\right)}\to \mathcal{Y}^{\left(k\right)}, k=1,2,\dots,K$. All input spaces have the same dimension, $\mathcal{X}^{\left(k\right)}\subseteq\mathbb{R}^{n}$, but the output spaces can be arbitrary, for example, discrete for classification and continuous for regression.

Let $L$ denote the number of layers formed by task-agnostic neurons, with the input layer being the first layer and the abstraction layer being the $L$-th layer. Let $I_{\left(l\right)}$ denote the set of neurons in the $l$-th layer, and $N_{\left(l\right)}$ the size of $I_{\left(l\right)}$. Each task-agnostic neuron represents a mapping $\Phi_{i}:\mathcal{X}\to\mathbb{R}$, where $\mathcal{X}=\bigcup\nolimits_{k=1}^{K}\mathcal{X}^{\left(k\right)}, i\in\bigcup\nolimits_{l=1}^{L}I_{\left(l\right)}$. Let $\Phi_{\left(l\right)}=\left[\Phi_{i}\right]_{i\in I_{\left(l\right)}}^{T}$ denote the concatenation of mappings of each layer, then $\Phi_{\left(l\right)}:\mathcal{X}\to\mathbb{R}^{N_{\left(l\right)}}$ can be defined recursively:
\vspace{-0.5 em}
\begin{align}
\Phi_{\left(1\right)}\left(x\right)&=x,\\
\Phi_{\left(l\right)}\left(x\right)&=\sigma_{\left(l\right)}\left(W_{\left(l-1\right)}\Phi_{\left(l-1\right)}\left(x\right)+b_{\left(l\right)}\right),\  l=2,\ 3,\ \dots,\ L,
\end{align}
\vspace{0.2 em}
where $W_{\left(l-1\right)}\in \mathbb{R}^{N_{\left(l\right)}\times N_{\left(l-1\right)}}$ is the weight matrix, $b_{\left(l\right)}\in \mathbb{R}^{N_{\left(l\right)}}$is the bias vector, and $\sigma_{\left(l\right)}: \mathbb{R}^{N_{\left(l\right)}}\to \mathbb{R}_{+}^{N_{\left(l\right)}}$ is the activation function. Note that $\Phi_{\left(l\right)}\left(x\right)$ is non-negative for $l\geq2$, and that the neuron $i$ is included in the feedforward process and thus "active" only if $\Phi_{i}\left(x\right)>0$. This means the network operates in a distributed manner, activating a portion of subnetworks and abstraction neurons for a given input. In other words, the network distributes its capacity over the input space.

In essence, ONE instantiates a special decomposition of $f^{\left(k\right)}$, which is $f^{\left(k\right)}=g^{\left(k\right)}\circ\Phi_{\left(L\right)}$, where $g^{\left(k\right)}$ represents the fully-connected layers in each task-specific group. If $\Phi_{\left(L\right)}$ represents really good abstractions, this decomposition can be highly efficient, because each $g^{\left(k\right)}$ will be quite simple and only need a few parameters. This means the majority of computation for each task is shared, and a new task merely adds a small increment to the whole network.

There are at least four differences between ONE and other neural networks (NNs): 1) \textbf{Partition structure:} NNs are mostly designed to implement specific tasks, making the entire network serve the tasks. 2) \textbf{Local parameter updating:} NNs’ parameters are fully updated during back propagation, which is one of reasons that causes knowledge  acquired in old tasks being forgotten in the process of implementing new tasks. In pre-trained models\cite{hinton2006reducing,mesnil2011unsupervised}, parameters are updated in a similar way to ours, but the knowledge is only a small part of all knowledge, which cannot be further accumulated in later learning. 3) \textbf{Abstraction locked layer:} The traditional NNs do not point out the specific form of abstractions, but only stay on the concept and description level. 4) \textbf{Dynamic network based on growth, competition and extinction:} The works of \citet{zhou2012online,philipp2017nonparametric,cortes2016adanet} explored dynamic neural networks, while none of them considered multi-task setting. The network\cite{xiao2014error} grows and branches only on the topmost layer, while ONE can change structures at any layer in the cognitive part.


\section{Constraint optimization formulation}
\vspace{-0.8 em}
The structure mentioned above is just a framework of abstraction generation, which satisfies reusability but not guarantee other properties of abstractions. So, we formulate the development of abstractions into a constrained optimization problem, where the constraints ensure the variety and simplicity of abstractions, and the objective function measures the effectiveness of abstractions.

\textbf{Variety constraint.}  ONE is intended for multiple tasks that can be radically different, so it must have the ability to generate and maintain various abstractions required to accomplish these tasks. Even if they are not required, a variety of abstractions would offer a wider perspective and probably lead to better solutions.

In ONE, each abstraction is represented by an abstraction neuron, which in turn corresponds to a subnetwork. Therefore, a variety of abstractions require a variety of subnetworks. The variety also provides an opportunity to search for specialized structures that express abstractions best.

Because the network capacity is fixed, in order to realize variety, we need to control the size of subnetworks and make sure that they do not overlap much. Accordingly, we introduce two constraints as followed.
\vspace{0 em}
\begin{align}
&A1:\quad \sum\nolimits_{j\in I_{\left(l\right)}}\mathbb{I}\left\{\Phi_{j}\left(x\right)>0\right\}\leq {V1}_{\left(l\right)},\ l=2,3,\ \dots,\ L-1,\ \forall x\in\mathcal{X},\\
&A2:\quad \sum\nolimits_{h\in I_{\left(l+1\right)}}\mathbb{I}\left\{W_{j\to h}\neq 0\right\}\leq {V2}_{\left(l\right)}^{\ j},\ \forall j\in I_{\left(l\right)},\ l=2,\ 3,\ \dots,\ L-1,\ \forall x\in\mathcal{X}.
\end{align}
\vspace{0 em}
Here, $\mathbb{I}\left\{\cdot\right\}$ is the indicator function. The first constraint limits the number of active neurons at each layer for any given input, thus controls the size of the activated subnetworks. The second constraint limits the number of outgoing connections of each neuron, thus reduces the overlap among subnetworks.

\textbf{Simplicity constraint.}  Once we develop good abstractions, we shall be able to grasp these concepts and principles, and use them to guide our actions and decisions. This greatly simplifies the tasks we encounter.

We introduce simplicity into ONE from two aspects. First, we require that only a few abstraction neurons are activated for any given input. This not only fosters simple explanation of the input, but also limits the size of task-specific subnetworks, enforcing simple decision making. Second, we expect abstractions to capture the commonality among entities. For clarity, we consider classification tasks only. We impose an upper bound on the total number of abstraction neurons that can be activated for all inputs belonging to the same category.

More concretely, let $\mathcal{C}^{\left(k\right)}$ denote the set of possible categories in the $k$-th classification task, and $\mathcal{X}^{c}$ denote the input space for each category $c\in \mathcal{C}^{\left(k\right)}$. We encourage simplicity by adding the following two constraints.
\vspace{0 em}
\begin{align}
&A3:\quad \sum\nolimits_{i\in I_{\left(L\right)}}\mathbb{I}\left\{\Phi_{i}\left(x\right)>0\right\}\leq S1,\ \forall x\in \mathcal{X},\\
&A4:\quad \sum\nolimits_{i\in I_{\left(L\right)}}\mathbb{I}\left\{ \sum\limits_{x\in\mathcal{X}^{c}}\mathbb{I}\left\{\Phi_{i}\left(x\right)>0\right\}>0 \right\}\leq S2,\ \forall c\in\mathcal{C}^{\left(k\right)},\ k=1,\ 2,\ \dots,\ K.
\end{align}
\vspace{0 em}
\textbf{Effectiveness constraint.}  Although abstractions are task-agnostic, some of them are particularly relevant to certain tasks while others are not. To boost task performance, we propose a matching process that selects highly effective abstraction neurons for each task. These selected neurons are then included in separate fully connected layers to adapt to task-specific details.

Thus, the network achieves effectiveness for tasks through two stages of optimization. First, it optimizes the selection variables $e_{i}^{\left(k\right)}$ for each abstraction neuron $i \in I_{\left(L\right)}$ and each task $k$. If neuron $i$ is selected for task $k$, $e_{i}^{\left(k\right)}=1$, and otherwise $e_{i}^{\left(k\right)}=0$. Second, the network optimizes the weights of the fully connected layers $W_{\left(\geq L\right)}^{\left(k\right)}$ to reduce task losses. For neurons with $e_{i}^{\left(k\right)}=0$, their connection weights to the $k$-th task-specific group $W_{i}^{\left(k\right)}$ are constrained to be zero.

We now define the objective function for optimizing $e_{i}^{\left(k\right)}$. For classification tasks, it is reasonable to assume that an abstraction neuron can promote task effectiveness if it has some preliminary ability to distinguish different categories. We evaluate the ability through the distribution of the neuron’s activations over inputs of all categories for any given task. If the neuron is activated by inputs of only a few categories, we expect that it has such ability. More formally, we optimize $e_{i}^{\left(k\right)}$ to select the $E$ neurons whose activation distributions have the lowest entropy.
\vspace{0 em}
\begin{equation}
\min\limits_{e_{i}^{\left(k\right)}}\sum\limits_{i\in I_{\left(L\right)}}e_{i}^{\left(k\right)}\cdot H_{i}^{\left(k\right)}
\end{equation}
\begin{equation}
\begin{aligned}
subject\ to \\
&A5:\quad \sum\nolimits_{i\in I_{\left(L\right)}}e_{i}^{\left(k\right)}=E,\ k=1,\ 2,\ \dots,\ K,\\
&A6:\quad W_i^{(k)}=0,\ if\ e_i^{(k)}=0,\ \forall i\in I_{(l)},\ k=1,\ 2,\ ...,\ K.
\end{aligned}
\end{equation}
\vspace{0 em}
Here
\vspace{0 em}
\begin{align}
&H_{i}^{\left(k\right)}=-\sum\nolimits_{c\in\mathcal{C}^{\left(k\right)}}P_{i}^{c}\cdot\log P_{i}^{c},\\
&P_{i}^{c}=\frac{\sum\nolimits_{x\in\mathcal{X}^{c}}\mathbb{I}\left\{\Phi_{i}\left(x\right)>0\right\}}{\sum\nolimits_{c\in\mathcal{C}^{\left(k\right)}}\sum\nolimits_{x\in\mathcal{X}^{c}}\mathbb{I}\left\{\Phi_{i}\left(x\right)>0\right\}}.
\end{align}
\vspace{0 em}
\textbf{Overall optimization problem.}  We now present the overall optimization problem that the network needs to solve. It integrates both the requirements for abstractions and the performance on each task.
\vspace{-0.5 em}
\begin{equation}
\min\limits_{W,e_{i}^{\left(k\right)}}\sum\limits_{k=1}^{K}\left(\mathcal{L}^{\left(k\right)}+\sum\limits_{i\in I_{\left(L\right)}}e_{i}^{\left(k\right)}\cdot H_{i}^{\left(k\right)}\right),
\end{equation}
\begin{equation}
\boxed{
subject\ to\ A1,\ A2,\ A3,\ A4,\ A5,\ A6.}
\end{equation}

Note that we do not include $\Phi_{i}$ in optimization variables, because they depend implicitly on the weights $W_{\left(<L\right)}$ between task-agnostic neurons. Note also that because $H_{i}^{\left(k\right)}$ depends on $\Phi_{i}$, the term $\sum\nolimits_{i\in I_{\left(L\right)}}e_{i}^{\left(k\right)}\cdot H_{i}^{\left(k\right)}$ is optimized jointly on $W$ and $e_{i}^{\left(k\right)}$. This implies that the differentiation ability of abstractions can also be improved.

\section{Network evolution algorithm}
\vspace{-0.8 em}
We observe that this optimization problem presents two major difficulties. First, the objective function includes discrete variables $e_{i}^{\left(k\right)}$. Also, the constraints depend on discrete values $\mathbb{I}\left\{\Phi_{j}\left(x\right)>0\right\}$ and $\mathbb{I}\left\{W_{j\to m}\neq 0\right\}$. Therefore, this difficulty of discreteness results from optimizing structure along with parameters, which is essential because the best structure for abstraction is unknown. The second difficulty comes from the nature of learning. In particular, tasks are not carried out simultaneously, but divided into batches and implemented sequentially. Consequently, abstractions for all tasks cannot be acquired at a time, but need to accumulate over time. This means the structure of the network keeps changing.

Standard gradient-based methods only work for continuous parameters and fixed structures, and thus are not suitable for our problem. Moreover, there is the same problem of optimizing discrete variables and dynamic structures in nature, which gets solved though natural evolution. Therefore, we propose an evolution-inspired optimization framework to search for good abstractions. More concretely, we identify three key elements of natural evolution: production, competition, and selection. Production generates a vast number of individuals, which then go through local competition. Nature selects the fittest individuals from the winners, and use them to guide a new round of production.

We incorporate these three components into our optimization framework to perform efficient parallel search of the structure space for good abstractions. In particular, we use production to grow new structures and maintain the variety abstractions. This corresponds to drawing a large sample from the structure space. These structures compete locally and only a small proportion of them get activated. This promotes the simplicity of abstractions, and improves search efficiency. The activated structures are then evaluated for their effectiveness and selected by each task. Only good structures survive, and they will affect the direction of production afterwards. We now describe each of these components in detail.

\begin{figure}
	\begin{minipage}[t]{0.45\textwidth}
		\centering
		\includegraphics[width=\linewidth]{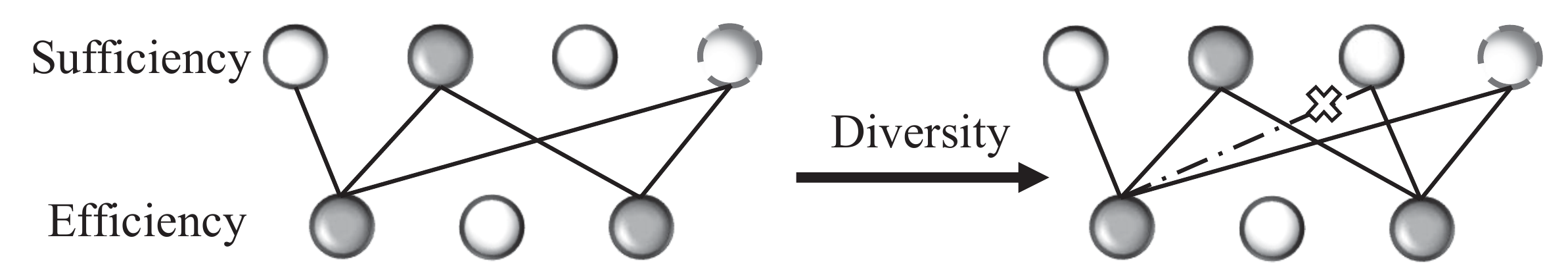}
		\caption[Production]{Production. The dark units repre
			-sent activated neurons, while the light ones
			do not. The dashed units denotes the newly
			activated neurons.}
		\label{fig:Production}
	\end{minipage}
	\quad
	\begin{minipage}[t]{0.45\textwidth}
		\centering
		\includegraphics[width=0.84\linewidth]{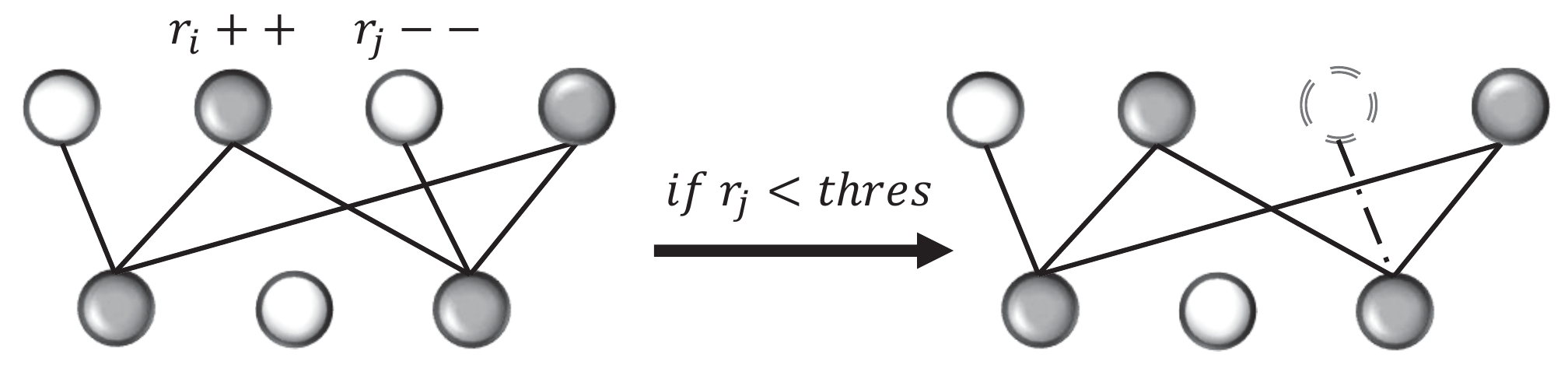}
		\caption[Competion]{Competition. The dark units denote winners, while the light ones denote losers. The dashed units indicate the dying neurons.}
		\label{fig:Competition}
	\end{minipage}
\end{figure}

\textbf{Production.} In our framework, production refers to growing connections among task-agnostic neurons. This is necessary for two reasons. First, production is a way of exploring the structure space. The more extensively we explore, the better structure we expect to find. Second, production enables the network to better adapt to new tasks which may require new structures (see Figure  \ref{fig:Production}).

We integrate production into the feedforward pass of the network, which means connections are created layer by layer. We design three production principles that help enhance the efficiency, sufficiency, and diversity of exploration. 1) Efficiency. In each feedforward pass, only the activated neurons can grow connections to the next layer. In this way, newly created structures are more likely to be activated and evaluated. Also, this helps capture regularities in the input data. 2) Sufficiency. To ensure that sufficient structures are explored, production is triggered if there are not enough activations in the next layer. 3) Diversity. The created structures should also be diverse. This is required by a variety of abstractions. We promote diversity by reducing the production probability of neurons which already have a large number of outgoing connections.


\begin{figure}
	\centering
	\begin{subfigure}[b]{0.45\textwidth}
		\centering
		\includegraphics[height=0.75in,width=1in]{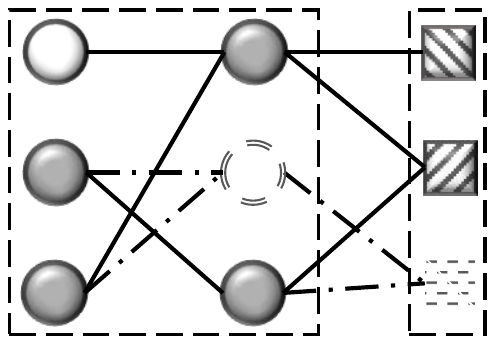}
		\subcaption{Death}
		\label{fig:Selection Death}
	\end{subfigure}\quad
	\begin{subfigure}[b]{0.45\textwidth}
		\centering
		\includegraphics[height=0.75in,width=1in]{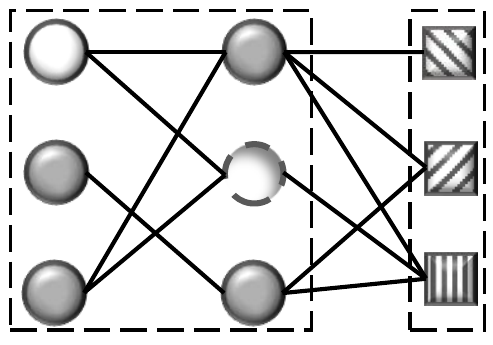}
		\subcaption{Growth}
		\label{fig:Selection Death}
	\end{subfigure}
	\caption[Selection]{(a) The third abstraction is not good enough and it dies along with the structure it is connected to. (b) The third one is new abstraction produced by a new structure.}
	\label{fig:Selection}
\end{figure}

\textbf{Competition.} We introduce competition among task-agnostic neurons within each layer. Specifically, for a given input, the neurons with the highest output values win the competition. Only the winners propagate their outputs to the next layer; other neurons are de-activated. This is similar in spirit to winner-take-all circuits which are common modules in the brain.

We believe that competition brings two benefits. First, it promotes the variety and simplicity of abstractions, because it controls the size of activated subnetworks and the number of activated abstractions. Second, it prunes unpromising structures on the fly – structures that contain deactivated neurons will not be evaluated and selected for the given task. This is essential to maintain search efficiency. Because the structure space is enormous and task-specific evaluation is costly, we can only afford to evaluate a very small proportion of all possible structures.

Since the network capacity is fixed, we need to give priority to promising structures and delete others if necessary. Therefore, we keep track of the promising structures by maintain a reward variable $r_{j}$ for each neuron $j$. Specifically, $r_{j}$ is incremented each time neuron $j$ wins a competition, and decremented when it loses.When $r_j$ is less than a threshold, the corresponding neuron dies (see Figure  \ref{fig:Competition}).


\textbf{Selection.} Selection aims to optimize the abstractions for task effectiveness. Because the objective function is non-differentiable, we take an iterative approach instead of standard gradient-based methods. At each iteration, we select the most effective abstractions from those that satisfy the variety and simplicity constraints. We preserve these abstractions and their corresponding structures, and eliminate others to allow new abstractions to be produced (see Figure  \ref{fig:Selection}). In order to make production more efficient, we guide the direction of production by giving advantage to components of effective structures in the production process.

Evaluating task effectiveness is costly, because computing the entropy of activation distributions requires a certain amount of input data. Consequently, compared to production and competition, selection operates on a larger time scale, and leads to significant modifications to the network structure.

The detailed algorithm for these three components is in the appendix.



\section{Experiments}
In all experiments of this section, we verify the feasibility of the model on the issues of single task and multiple consecutive tasks respectively. The prediction accuracy and effectiveness of the model are the main concerns, andwe verify effectiveness in terms of the forgotten rate of knowledge and the sharing rate of abstractions. In order to evaluate the performance of ONE more clearly, no special strategies, such as augmenting data with transformations, dropout or noise, were used. We did not fine-tune any parameters such as parameters initialization in the abstraction layer, either. Thus, our objective is to evaluate the feasibility of the model rather than achieving the absolute best testing scores. Moreover, it is well known that one of the drawbacks of evolution algorithms is that the speed of convergence is slow, so we only validate the feasibility on MNIST dataset. In the future, we will verify our ideas on more datasets.

The MNIST dataset consists of 60K training images and 10K test images of handwritten digits from 0 to 9, where the size of each image is $28\times 28$. In the following experiments, we used all available training data to train the model. Furthermore, we divided the whole classification task into 5 tasks, each of which is a binary classification task on each class.

Our model introduces several hyper-parameters , which we refer to as extinction and growth parameters, in the process of generating abstractions. In terms of growth parameters, parameter $a_l$  controls the minimum number of neurons that each training sample activates on layer $l$, parameter $b_l$ indicates that each neuron in layer $l$ can be connected to $b_l$ neurons in layer $l-1$ at most, parameter $c_l$ controls the number of neurons that win in layer $l$ by directly enforcing a winner-take-all sparsity constraint and coordinates with $a_l$  to produce good neurons. In terms of extinction parameters, parameter $d$ and parameter $e$ jointly indicate that when the activation number of a neuron is greater than $d$ and the activation number of synapses connected with the neuron is less than $e$, then the synaptic dies. And if a neuron does not have any outgoing connection, the neuron dies. If growth parameters are set too large, they will promote the exploration of the structure, so as to find the optimal structure, but it is  easy to cause the redundancy of the structure. On the contrary, if growth parameters are set too small, the structure exploration is slow and the number of shared neurons are reduced, which is not conducive to the generation of abstractions. What’s more, too large extinction parameters will not be conducive to the accumulation of abstractions, while too small parameters lead to abstraction redundancy. Instead of using the 2D structure of the images, we turn each image into a vector of pixels, where the pixel values were rescaled to $[0,\ 1]$. According to the general configuration of CNNs, we set the parameters as follows: the configuration of layers in the cognitive part: $\left\{784,\ 500,\ 500,\ 500\right\}$, the configuration of layers in the decision part: $\left\{200,\ 50,\ 2\right\}$, $a=\left\{200,\ 160,\ 100\right\}$, $b=\left\{50,\ 50,\ 50\right\}$, $c=\left\{100,\ 80,\ 60\right\}$, $d=3000$, $e=300$.

\begin{figure*}[t]
\begin{minipage}{\linewidth}
	\centering
	\begin{minipage}[c]{0.56\linewidth}
		\captionsetup{type=figure} 
		\centering
		\subcaptionbox{Abstraction generation\label{fig:Abstraction generation}}
        [0.49\linewidth] 	{\includegraphics[height=0.95in]{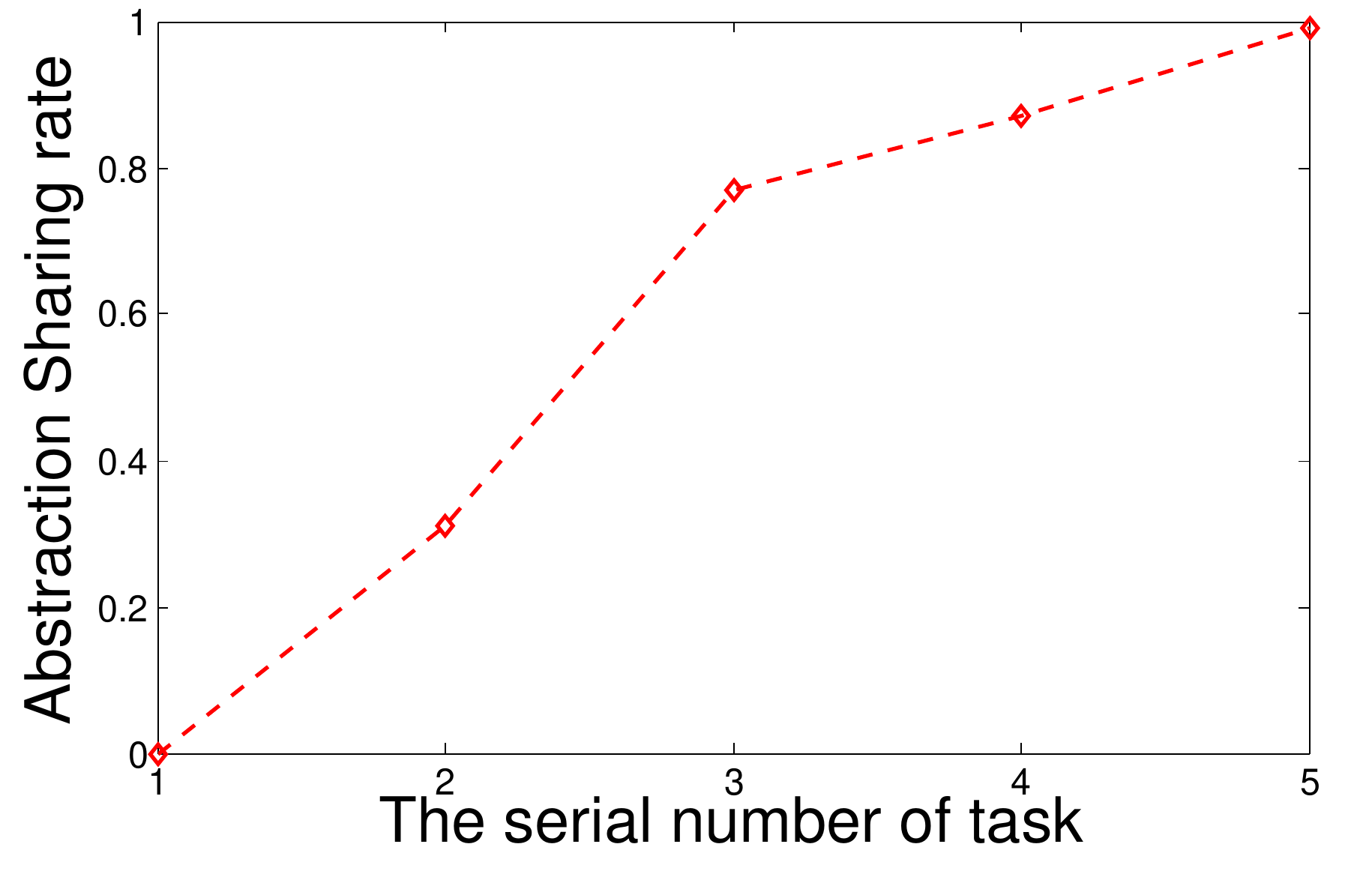}}
		\subcaptionbox{Abstraction sharing\label{fig:Abstraction sharing}}
        [0.49\linewidth]	{\includegraphics[height=0.95in]{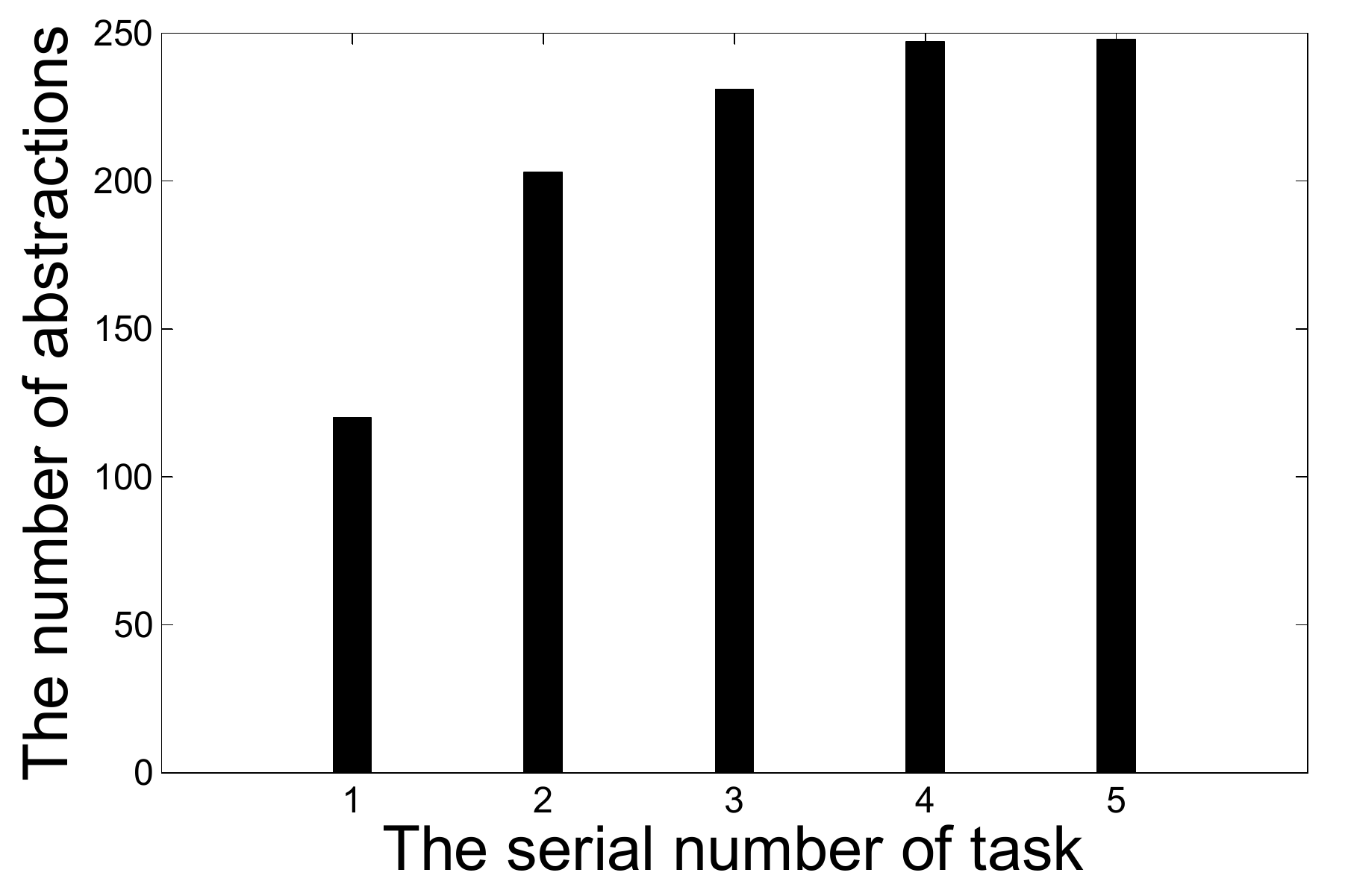}}
		\caption{(a) shows the proportion of abstractions generated by the previous tasks in the total abstractions used by the current task. (b) shows the total number of abstractions as tasks increase.}
		\label{fig:Abstraction generation and sharing}
	\end{minipage}
	\quad
	\begin{minipage}[c]{0.4\linewidth}
		\centering
		\captionof{table}{Test accuracy on MNIST without data augmentation.}
		\label{table:Test accuracy}
		\resizebox{1.0\linewidth}{!}
		{\begin{tabular}{@{}ccccc@{}}
			\toprule[1.0pt]
			& Subtask & Single Task & Consecutive tasks\\
            \midrule[0.5pt]
			& 1     & 99.81\%       & 99.81\%     \\
			& 2     & 97.06\%       & 97.01\%      \\
            & 3     & 98.83\%       & 98.67\%  \\
            & 4     & 99.04\%       & 99.45\%  \\
            & 5     & 96.72\%       & 97.23\%  \\
            \bottomrule[1.0pt]
		\end{tabular}}
	\end{minipage}
\end{minipage}
\end{figure*}

The second column of Table \ref{table:Test accuracy} shows the experimental results in five tasks. ONE obtains pretty good test scores, which proves the feasibility of the model. The reason why the recognition rates for task 2 and 5 are not as good as those for the rest 3 tasks may be that the difference between the digits is small.


In incremental learning, tasks come in sequence. All parameters in the cognitive part are shared, including structure, weight and bias, while the parameters in the decision part for each task are not shared. The third column of Table \ref{table:Test accuracy} shows the test accuracy of each task in incremental learning. Our model, ONE, performs almost the same as these batch models in single-task experiments, and even outperforms them in the last two tasks. Figure \ref{fig:Abstraction sharing} shows that ONE performs each task based on part of the total abstraction (500), and thus activates only a small proportion of the whole network for each task. To go further, we count the abstraction sharing rate between current tasks and all previous tasks (see figure \ref{fig:Abstraction generation}). The result shows that fewer new abstractions are produced as tasks increase, which means that abstractions produced by ONE are highly reusable. This reduces energy consumption and boosts learning speed. And the reason why the test accuracy increases in incremental learning is that some abstractions produced in other tasks are useful in current task, while they cannot be generated by current data. Moreover, because of the accumulation of the abstractions and the non-overlap of the decision part for different tasks, ONE does not forget, and thus can learn continually. We believe that abstractions play an extremely important role in lifelong learning, and the advantages of ONE will be more prominent as the number of tasks increases.

\section{Conclusion}
We proposed ONE, a completely new framework for learning abstractions and achieving human-like intelligence. The partition structure provides the foundation for abstraction accumulation across tasks, enabling knowledge transfer and lifelong learning. The constrained optimization formulation directly specifies the properties of abstractions, turning abstraction learning into a well-defined problem. The network evolution algorithm effectively searches through the structure space, ensuring the quality of abstractions.

While human beings may represent and learn abstractions in a different way, we have demonstrated that the abstractions learned by ONE can offer similar benefits as those learned by human beings. In the future, we plan to extend the experiments to larger datasets and perform cross-domain abstraction learning. Another interesting direction is to learn abstractions of tasks. This would allow sharing among task-specific parts, and also improve interpretability of decision making.

\bibliographystyle{plainnat}  
\bibliography{reference}

\end{document}